\documentclass[11pt]{article}
\usepackage[margin=1in]{geometry}
\usepackage[T1]{fontenc}
\usepackage{graphicx}
\usepackage{amsmath,amssymb}
\usepackage{booktabs}
\usepackage{multirow}
\usepackage[table]{xcolor}
\usepackage{microtype}
\usepackage{tikz}
\usetikzlibrary{arrows.meta, positioning, calc, fit, backgrounds, shapes.geometric, decorations.pathreplacing, patterns, shadows}
\usepackage[hidelinks]{hyperref}
\usepackage{authblk}
\hypersetup{pdftitle={The Diagnosis a Reporter Leaves Unspoken: Surfacing Frozen Tumor Features for
Brain-Tumor MRI Reporting},
pdfauthor={Khawaja Murad ul Hassan, Ruqiyya Adil, Adil Qayyum, Rida Hassan, Asad Mansoor Khan,
Muhammad Usman Akram, Mehran Ebrahimi},
pdfsubject={MLCN 2026}}

\newcommand{\xgrammar}{\textsc{XGrammar}}
\newcommand{\sys}{NeuroFusion}
\newcommand{\dxpin}{\textsc{Dx-Pin}}
\newcommand{\mednext}{MedNeXt-L}
\definecolor{nfBlue}{HTML}{C3DCF2}
\definecolor{nfBlueE}{HTML}{2C5AA0}
\definecolor{nfOrange}{HTML}{FBE0C2}
\definecolor{nfOrangeE}{HTML}{D97706}
\definecolor{nfGray}{HTML}{D9DDE2}
\definecolor{nfGrayE}{HTML}{4B5563}
\definecolor{cWrong}{HTML}{C62828}
\definecolor{cVague}{HTML}{1F6FB2}
\definecolor{cOmit}{HTML}{2E7D32}

\title{The Diagnosis a Reporter Leaves Unspoken: Surfacing Frozen Tumor Features for Brain-Tumor MRI Reporting\thanks{Accepted at MLCN 2026, a workshop held in conjunction with MICCAI 2026. The Version of Record will appear in \emph{Medical Image Computing and Computer Assisted Intervention -- MICCAI 2026 Satellite Events}, Springer Lecture Notes in Computer Science. This preprint is the authors' own version, typeset independently of the publisher.}}

\author[1]{Khawaja Murad ul Hassan\thanks{Corresponding author: \texttt{khawajamurad@outlook.com}}}
\author[1]{Ruqiyya Adil}
\author[2]{Adil Qayyum}
\author[3]{Rida Hassan}
\author[1]{Asad Mansoor Khan}
\author[1]{Muhammad Usman Akram}
\author[4]{Mehran Ebrahimi}
\affil[1]{National University of Sciences and Technology, Islamabad, Pakistan}
\affil[2]{Consultant Radiologist, Rawalpindi, Pakistan}
\affil[3]{Bahria University Health Sciences Campus, Islamabad, Pakistan}
\affil[4]{Faculty of Science, Ontario Tech University, Oshawa, ON, Canada}
\date{}

\begin{document}
\maketitle

\begin{abstract}
\noindent A capable brain-MRI report generator can still be, in effect, diagnostically silent. When a
multi-chain chain-of-thought (CoT) reporter built on a medical Mistral-7B backbone is evaluated on
held-out cohorts, it names most meningiomas and almost all metastases ``glioma'' (diagnosis recall
$0.44$/$0.07$). Yet the answer is not absent from the model: a supervised linear probe applied to its
\emph{frozen} segmentation features recovers the three tumour cohorts at $0.82$ macro-F$_1$ ($5$-fold
cross-validation; chance $\approx\!0.33$). We introduce \sys{}, an assistive reporter that \emph{surfaces}
this latent signal rather than overriding it: discriminative field-classifier heads over per-lesion
features condition a fast, single-pass draft-then-review decoder on their committed outputs. Built on the
\emph{identical} Mistral backbone, this restores the diagnosis (meningioma $0.92$, metastasis $0.75$) and
wins \textbf{$8$ of $9$} prose-content comparisons across three held-out cohorts (RaTEScore, RadGraph-F$_1$,
GREEN; Holm-corrected paired BCa), with no significant loss on the ninth, at \textbf{$5$-$6\times$ lower
latency} ($\approx\!80$ vs.\ $457$\,s/case). A controlled \emph{negative result} sharpens the mechanism: a
learned \emph{diagnosis pin} that overrides the decoder instead of merely informing it collapses
out-of-distribution metastasis recall to $0.03$. Grammar-constrained decoding keeps $92.3\%$ of records
schema-valid, making every sentence entailment-checkable ($7.5\%$ contradicted vs.\ $36.8\%$ for the direct
baseline). In a blinded nine-case pilot, two board-certified neurologists independently rated \sys{} highest
in every tumour type, the only system with zero critical errors, and gave it the top-rated sign-off in eight
of nine cases (six outright, two ties).

\end{abstract}

\section{Introduction}
\label{sec:intro}
Brain tumours are both high-stakes and common worldwide: GLOBOCAN recorded $321{,}731$ new malignant brain
and CNS cancer cases in $2022$, with $248{,}500$ deaths~\cite{globocan2022}, and MRI remains the front-line
imaging modality. Access, however, is deeply uneven: roughly two-thirds of the world's population has no
access to \emph{any} diagnostic imaging~\cite{hricak2021lancet,mariani2017imaging}. Where imaging is
available, radiologists face a punishing workload -- reading an image roughly every $3$-$4$\,s~\cite{mcdonald2015workload},
with a single brain MRI taking about $11$ minutes to read and dictate, and over a third of radiologists
reporting burnout~\cite{ashraf2023burnout,alyassin2018time}. There is evidence automation can help:
keyword-based AI drafting alone has been shown to cut intracranial-tumour reporting time by roughly
$28\%$ with no loss of quality~\cite{dong2025aireport}. We focus on the most automatable slice of this
workflow -- structured findings plus a first-pass narrative -- and target a record that, unlike existing
generators, is machine-checkable and commits to a diagnosis instead of hedging.

Existing tools fall short in complementary ways. Fluent 3D medical vision-language models and brain report
generators expose no checkable schema and default to the most common diagnosis, while a rule that reads
fields directly off a segmentation mask fills a schema but offers only a fixed differential and is capped
by whatever the mask shows.

The omission is not simply a perception failure. On our held-out cohorts, a strong chain-of-thought (CoT)
reporter sharing our backbone misreads more than half of meningiomas and \emph{nearly every} metastasis as
``glioma'' (diagnosis recall $0.44$/$0.07$); yet a \emph{supervised} linear probe over the same model's
\emph{frozen} segmentation bottleneck separates the three cohorts at $0.82$ macro-F$_1$ ($5$-fold
cross-validation, chance $\approx\!0.33$; per-cohort $0.85$/$0.88$/$0.75$). The diagnosis, in other words,
is linearly accessible in the model's own features but never verbalized -- a phenomenon we call
\emph{diagnostic suppression}. \sys{} closes this gap through \emph{head-surfacing}: the same backbone runs
discriminative field-classifier heads over the per-lesion features and conditions a fast, single-pass
draft-then-review decoder on their committed outputs, restoring the diagnosis (meningioma $0.92$, metastasis
$0.75$) without introducing a separate diagnosis model. Against the identical backbone's CoT variant it wins
$8$ of $9$ prose-content comparisons with no significant regression on the ninth, at $5$-$6\times$ lower
latency, and the gain is individually significant on the held-out metastasis cohort
(Sec.~\ref{sec:crossdata}); a controlled \emph{diagnosis-pin} negative result further sharpens the
mechanism. \sys{} also adds an auditability layer no baseline VLM offers -- per-sentence faithfulness
checking over its machine-checkable record (Sec.~\ref{sec:analysis}) -- and in a blinded reader study,
neurologists rate it highest in every tumour type (Table~\ref{tab:reader}).

\noindent\textbf{Contributions:} (i) a \emph{diagnostic-suppression} phenomenon, evidenced by the gap
between a $0.82$-F$_1$ frozen-feature probe and a same-backbone CoT reporter's $0.44$/$0.07$ diagnosis
recall; (ii) \emph{discriminative-head conditioning} as a remedy, yielding Holm-significant same-base gains
on $8$ of $9$ prose-content comparisons at $5$-$6\times$ lower latency, individually significant on the
held-out metastasis cohort (Sec.~\ref{sec:crossdata}); (iii) a controlled \emph{diagnosis-pin} negative
result; (iv) a verifiable audit layer evaluated under a pre-specified statistical protocol
(Holm/TOST/BCa, Sec.~\ref{sec:analysis}); and (v), to our knowledge, the first blinded neurologist reader
study for this task, in which \sys{} is rated highest in every tumour type (Table~\ref{tab:reader}).

\section{Related Work}
\label{sec:related}
\textbf{Brain MRI reporting.} AutoRG-Brain~\cite{autorgbrain} pairs anomaly segmentation with a
visually-prompted language model for findings generation; MAIRA-2~\cite{maira2} grounds reporting in image
regions; BrainGemma3D~\cite{braingemma3d} is a single-sequence 3D generator; concurrent work targets 3D
brain-tumor reporting more broadly (Brain3D~\cite{brain3d}, RadFM~\cite{radfm}). All of these ground reports
in \emph{image regions} but leave the diagnosis itself to the language model and expose no checkable
schema; we benchmark directly against AutoRG-Brain and BrainGemma3D. \textbf{3D medical VLMs.}
M3D-LaMed~\cite{m3dlamed} and LLaVA-Med~\cite{li2023llavamed} pair a vision encoder with an LLM to produce
free text but target no schema at all. \textbf{Constrained decoding and faithfulness.}
\xgrammar{}~\cite{xgrammar2024} guarantees syntactic, not semantic, validity; entity-level metrics such as
RadGraph~\cite{radgraph,delbrouck2024radgraphxl}, RaTEScore~\cite{zhao2024ratescore}, and
GREEN~\cite{ostmeier2024green} assess grounding beyond simple $n$-gram overlap. \textbf{Label-conditioned
reporting.} Conditioning a generator on predicted labels is well established in chest X-ray reporting
(tag-conditioned~\cite{jing2018automatic}; CheXpert-label-conditioned~\cite{liu2019clinically});
region-grounding, by contrast, conditions on \emph{where} a finding is rather than \emph{what} it is.
\textbf{Positioning.} Our contribution is not head-conditioning as such, but the finding that the gain
requires a decoder \emph{trained to commit} to the fields -- mere availability is not enough: a same-base
CoT model handed the identical head argmax as text still misreads the diagnosis. We hard-set the enum
fields but deliberately leave the diagnosis itself to the language model.

\section{Method}
\label{sec:method}
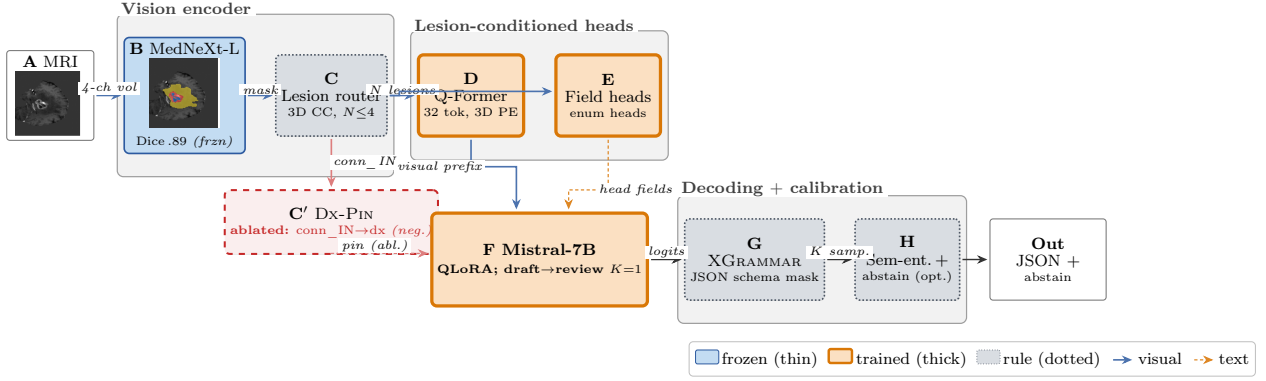
\begin{figure}[tb]
\centering
\makebox[\textwidth][c]{\resizebox{\textwidth}{!}{%
\begin{tikzpicture}[
    >={Stealth[length=1.8mm,width=1.5mm]}, font=\footnotesize,
    every node/.style={font=\footnotesize},
    stage/.style ={rounded corners=2.5pt, draw, line width=0.8pt, minimum width=18mm,
                   minimum height=14mm, align=center, inner sep=2.6pt, font=\scriptsize,
                   drop shadow={shadow xshift=0.3pt, shadow yshift=-0.4pt, opacity=0.16, fill=black!70}},
    frozen/.style ={stage, fill=nfBlue,   draw=nfBlueE},
    trained/.style={stage, fill=nfOrange, draw=nfOrangeE, line width=1.6pt},
    rule/.style   ={stage, fill=nfGray,   draw=nfGrayE, densely dotted},
    LMbox/.style  ={trained, minimum width=30mm, minimum height=16mm, font=\scriptsize\bfseries},
    inout/.style  ={rounded corners=1.5pt, draw=black!55, line width=0.5pt, minimum width=17mm,
                    minimum height=14mm, align=center, inner sep=2.6pt, font=\scriptsize, fill=white},
    flow/.style    ={->, line width=0.7pt, draw=black!80, shorten >=1pt, shorten <=1pt},
    visflow/.style ={->, line width=0.8pt, draw=nfBlueE!85, shorten >=1pt, shorten <=1pt},
    txtflow/.style ={->, line width=0.6pt, dash pattern=on 1.6pt off 1.2pt,
                     draw=nfOrangeE!88, shorten >=1pt, shorten <=1pt},
    edge_lbl/.style={font=\scriptsize\itshape, fill=white, inner xsep=1.5pt, inner ysep=0.5pt},
    grp/.style     ={draw=black!38, fill=black!5, rounded corners=3pt, inner sep=4pt, line width=0.6pt},
    grplbl/.style  ={font=\scriptsize\bfseries, text=black!75, anchor=south west, fill=white,
                     inner xsep=2pt, inner ysep=0.5pt},
  ]
  \def\colSep{5mm}
  \def\rowGap{20mm}
  \node[inout, minimum width=15mm]   (in)  at (0,0)
       {\textbf{A}~MRI\\[1pt]\includegraphics[width=12mm]{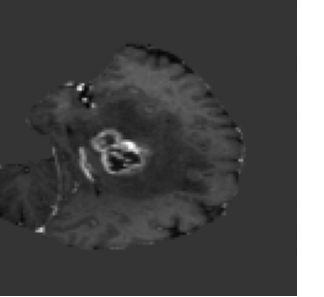}};
  \node[frozen, right=\colSep of in, minimum width=15mm] (seg)
       {\textbf{B}~\mednext\\[1pt]\includegraphics[width=12mm]{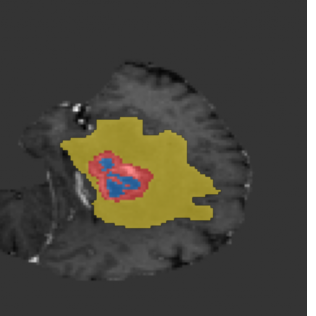}\\[-1pt]\tiny Dice\,$.89$ \emph{(frzn)}};
  \node[rule,   right=\colSep of seg](rt)
       {\textbf{C}\\[-1pt]Lesion router\\[-2pt]\tiny 3D CC, $N{\leq}4$};
  \node[trained, right=\colSep of rt](qf)
       {\textbf{D}\\[-1pt]Q-Former\\[-2pt]\tiny $32$ tok, 3D PE};
  \node[trained, right=\colSep of qf](fc)
       {\textbf{E}\\[-1pt]Field heads\\[-2pt]\tiny enum heads};
  \node[stage, dashed, draw=cWrong, fill=cWrong!8, line width=1.0pt, below=9mm of rt, minimum width=21mm, minimum height=11mm] (dx)
       {\textbf{C$'$}~\dxpin{}\\[-2pt]\tiny {\color{cWrong}\textbf{ablated:} conn\_IN$\to$dx \emph{(neg.)}}};
  \node[LMbox, below=\rowGap of $(qf)!0.5!(fc)$] (lm)
       {\textbf{F}~Mistral-7B\\[-1pt]\tiny QLoRA; draft$\to$review $K{=}1$};
  \node[rule,  right=\colSep+1mm of lm] (xg)
       {\textbf{G}\\[-1pt]\xgrammar{}\\[-2pt]\tiny JSON schema mask};
  \node[rule,  right=\colSep of xg]     (ab)
       {\textbf{H}\\[-1pt]Sem-ent.\,$+$\\[-2pt]\tiny abstain (opt.)};
  \node[inout, right=\colSep of ab, minimum width=20mm] (out)
       {\textbf{Out}\\[-1pt]JSON $+$\\[-2pt]\tiny abstain};
  \draw[visflow] (in)  -- node[edge_lbl, above]{\tiny 4-ch vol} (seg);
  \draw[visflow] (seg) -- node[edge_lbl, above]{\tiny mask} (rt);
  \draw[visflow] (rt)  -- node[edge_lbl, above]{\tiny $N$ lesions} (qf);
  \draw[visflow] (rt.east) -- ++(2mm,0) |- ([yshift=2pt]fc.west);
  \draw[visflow, draw=cWrong!60] (rt.south) -- (dx.north) node[edge_lbl, midway, right=-1pt]{\tiny conn\_IN};
  \draw[txtflow, draw=cWrong!60] (dx.south) |- ([yshift=3pt]lm.west) node[edge_lbl, pos=0.72, above]{\tiny pin (abl.)};
  \draw[visflow] (qf.south) -- ++(0,-5mm) -| ([xshift=-4mm]lm.north)
                 node[edge_lbl, pos=0.16, left=0pt]{\tiny visual prefix};
  \draw[txtflow] (fc.south) -- ++(0,-9mm) -| ([xshift=5mm]lm.north)
                 node[edge_lbl, pos=0.16, right=0pt]{\tiny head fields};
  \draw[flow] (lm)  -- node[edge_lbl, above]{\tiny logits} (xg);
  \draw[flow] (xg)  -- node[edge_lbl, above]{\tiny $K$ samp.} (ab);
  \draw[flow] (ab)  -- (out);
  \begin{scope}[on background layer]
    \node[grp, fit=(seg)(rt), inner ysep=11pt, inner xsep=3pt] (gV) {};
    \node[grp, fit=(qf)(fc),  inner ysep=11pt, inner xsep=3pt] (gH) {};
    \node[grp, fit=(xg)(ab),  inner ysep=11pt, inner xsep=3pt] (gD) {};
  \end{scope}
  \node[grplbl] at ([yshift=-1pt]gV.north west) {Vision encoder};
  \node[grplbl] at ([yshift=-1pt]gH.north west) {Lesion-conditioned heads};
  \node[grplbl] at ([yshift=-1pt]gD.north west) {Decoding $+$ calibration};
  \node[draw=black!25, fill=white, rounded corners=2pt, inner sep=3pt, line width=0.35pt,
        anchor=north west, font=\scriptsize] at ([xshift=2mm,yshift=-3mm]gD.south west)
        {\begin{tabular}{@{}c@{\,}l@{\;\;}c@{\,}l@{\;\;}c@{\,}l@{\;\;}c@{\,}l@{\;\;}c@{\,}l@{}}
            \tikz\node[fill=nfBlue, draw=nfBlueE, rounded corners=1pt, minimum width=4mm,
                       minimum height=2.4mm, line width=0.3pt]{}; & frozen (thin)
          & \tikz\node[fill=nfOrange, draw=nfOrangeE, rounded corners=1pt, minimum width=4mm,
                       minimum height=2.4mm, line width=1.2pt]{}; & trained (thick)
          & \tikz\node[fill=nfGray, draw=nfGrayE, rounded corners=1pt, minimum width=4mm,
                       minimum height=2.4mm, line width=0.3pt, densely dotted]{}; & rule (dotted)
          & \tikz\draw[visflow,->] (0,0)--(4mm,0); & visual
          & \tikz\draw[txtflow,->] (0,0)--(4mm,0); & text
          \end{tabular}};
\end{tikzpicture}}}
\caption{\sys{} architecture. A frozen \mednext{} backbone (B) segments the tumor; a connected-component
router (C) turns the $N{\leq}4$ retained lesions into per-lesion queries over the shared feature map
(volume-ranked, centroid-encoded) for a Q-Former (D) and field-classifier heads (E). Visual tokens and the
\emph{surfaced head outputs} condition the QLoRA-tuned Mistral-7B (F) in a single draft-then-review pass
($K{=}1$); an \xgrammar{} mask (G) enforces schema-valid JSON and an optional abstention gate (H) emits the
report. The diagnosis pin (C$'$, labelled \dxpin{}, dashed) is tested and dropped
(Sec.~\ref{sec:crossdata}).}
\label{fig:pipeline}
\end{figure}

\paragraph{Pipeline overview.}
\sys{} turns a four-sequence MRI volume into a single structured record in one decoder pass. A frozen
MedNeXt~\cite{roy2023mednext} backbone, pretrained across multiple cohorts (validation mean foreground Dice
$0.89$), segments the tumour; connected-component analysis yields per-lesion masks, and a router keeps only
the four largest lesions. Each lesion's pooled features -- tagged with a factorized 3D positional encoding
recording where in the volume it sits -- are compressed by a per-lesion Q-Former~\cite{li2023blip2}: a set
of learned query vectors that cross-attend the features and emit a fixed $32$ tokens per lesion, so that a
scan with any number of lesions reaches the language model as a bounded, volume-ordered visual prefix.
These tokens, together with the \emph{surfaced field-head outputs}, condition a
QLoRA~\cite{dettmers2023qlora}-tuned medical LLM that decodes the report under a schema grammar
(Fig.~\ref{fig:pipeline}). The language model is LLaVA-Med~v1.5's medical
Mistral-7B~\cite{li2023llavamed} -- the \emph{same} model used by the LLaVA-Med baseline, so any gain
reflects the architecture rather than a stronger backbone.

\paragraph{Head-surfacing and single-pass decoding.}
The output record is JSON: a fixed set of categorical fields (composition, enhancement pattern, mass
effect, and others), a differential diagnosis, and free-text findings and impression. Head-surfacing means
linear enum classifiers over the pooled per-lesion Q-Former features predict the structured fields
directly; the diagnosis itself is deliberately \emph{not} a head -- it is left to the language model to
recover. The heads' argmax (together with location inferred from segmentation geometry) is injected into
the prompt as text (e.g.\ ``temporal; heterogeneous; ring'') rather than left for the LM to draft on its
own, which tends to over-generate lesions and misstate the diagnosis; cued this way, the LM recovers both
per-field accuracy \emph{and} the diagnosis (Sec.~\ref{sec:crossdata}). Decoding uses \xgrammar{}, masking
the logits at each step to only the tokens the schema still permits, so a syntactically invalid record
cannot be produced. The whole process is a single \emph{draft-then-review} pass ($K{=}1$, versus the CoT
baseline's $K{=}2$ chains times four sub-questions): the model samples the record once, and the draft's
enum fields are then reconciled against the head argmax (never the diagnosis) within that same pass, rather
than through a second LM call -- replacing a slower CoT pipeline for no loss in quality. Only the router,
Q-Former, field classifiers, and LoRA adapter are trained; a \emph{record-verbalizer} variant (adapter
disabled) is used to maximize faithfulness (Sec.~\ref{sec:analysis}).

\section{Experimental Setup}
\label{sec:setup}
\paragraph{Data.} Training uses BraTS-2020~\cite{menze2015brats}; $121$ of its cases carry human-authored,
radiologist-reviewed structured reports (diagnosis drawn from histopathology or differential; $98\%$ pass
validation), from which we hold out $39$ in-distribution test cases and $50$ calibration cases. Two further,
in-distribution RadGenome cohorts -- glioma (GLI, $n{=}60$; $n{=}54$ once subject-twins are excluded) and
meningioma (MEN, $n{=}60$) -- extend training in a patient-disjoint way (held out, though not zero-shot),
while BraTS-MET ($n{=}60$) is held out specifically for the \emph{reporter}: the Q-Former, field heads, and
LoRA decoder never see a single metastasis report during training (the segmenter is metastasis-fine-tuned,
and the metastasis references come from AutoRG-Brain's corpus, so the reporter itself is out-of-distribution
even though the segmenter is not). All cohorts are deduplicated at the level of the same subject, with a
preflight gate confirming zero train/test overlap by both patient identity and content hash; six glioma
subject-twins are dropped for the twin-excluded contrast ($n{=}54$; the external comparison retains the
full $n{=}60$). \emph{Data use:} every volume comes from a public dataset under its original license; no new
patient data were collected, and no ethics approval beyond what the source datasets already required was
needed. The $121$ structured reports, authored by the study team, will be released publicly once curation
is complete.

\paragraph{Systems and baselines.} The primary comparison is \emph{same-base}: all systems share the
\emph{identical} Mistral LM. We compare \sys{} at two operating points (draft-then-review;
record-\textbf{verbalizer}) against the prior \textbf{CoT} reporter (multi-chain, $K{=}2$) and against a
\textbf{$+$diagnosis-pin} variant (a negative result, see below). The same-base CoT is given the identical
head argmax and confidence values as text, and its diagnosis recall remains $0.44$/$0.07$ regardless.
External baselines are M3D-LaMed and LLaVA-Med (\sys{}'s own LM) evaluated on their native free-text task,
with fields extracted by a judge model; AutoRG-Brain~\cite{autorgbrain}, run on the same predicted mask
(never its own benchmark-trained segmenter); and BrainGemma3D~\cite{braingemma3d} (meningioma and metastasis
only, since its training overlaps our glioma cohort).

\paragraph{Metrics and protocol.} We report schema validity; a per-field \emph{union-class macro-F$_1$ with
an over-prediction penalty} (computed on the predicted segmentation, best-of-$K$); three entity-level
narrative metrics scored prose-versus-prose (RadGraph-F$_1$~\cite{delbrouck2024radgraphxl},
RaTEScore~\cite{zhao2024ratescore}, GREEN~\cite{ostmeier2024green}); and a blinded $1$-$5$ clinical rubric
across five dimensions -- lesion identification, characterization, mass effect, differential diagnosis, and
readability (mean $=$ Clin-O; judged by Claude Opus~4.8). Diagnosis recall (whether the top differential
names the correct cohort type) is reported as a \emph{descriptive} lexical proxy, not a primary endpoint
(Fig.~\ref{fig:reports}). All statistical comparisons are \emph{pre-specified}: the primary endpoint is
RaTEScore on meningioma; the superiority family is the nine same-base cells $\{$RaTEScore, RadGraph,
GREEN$\}\times\{$GLI, MEN, MET$\}$, Holm-corrected, with TOST equivalence bounds of $\pm0.03$ (under half
the metric's standard deviation); the external comparison is the pre-specified $12$-test metastasis family
$\{$RaTEScore, GREEN, Clin-O$\}$ against each of the four external baselines. All intervals are two-sided
BCa bootstraps over paired per-case differences ($n_{\text{boot}}{=}20000$); the reported intervals come
from a single resampling seed, and five additional seeds confirm the sign of every bound. We never report a
directionally-non-significant cell as a win.

\paragraph{Reader study.} Two board-certified neurologists \emph{independently} rated de-identified reports
(behind opaque labels A to F, in a fixed order) for nine held-out cases ($3$ per cohort, selected at random
\emph{before} any scoring took place), across five $0$-$5$ axes, a \emph{critical-error} flag, and a forced
sign-off (we report the mean of the two raters). As a blinded \emph{pilot} we report descriptive means and
critical-error counts rather than $p$-values (Sec.~\ref{sec:reader}).

\section{Results: in distribution and the cross-cohort win}
\label{sec:results}\label{sec:crossdata}
\paragraph{In distribution ($n{=}39$, BraTS-2020).}
\sys{} emits schema-valid records for $92.3\%$ of cases ($36/39$; Wilson $95\%$ CI $[0.80,0.97]$), a rate
that collapses toward zero without grammar-constrained decoding. At its record-verbalizer operating point it
reaches an Opus clinical score of $3.11$ and a RadGraph score of $0.275$, and beats both VLM baselines on
structured-field accuracy ($0.449$ vs.\ $\le\!0.379$); head-surfacing's main payoff, however, shows up
cross-cohort.

\paragraph{Cross-cohort: $8$ of $9$ prose-content wins, no losses.} On the larger cohorts
(Table~\ref{tab:crossdata}), \sys{} beats the same-base CoT on \emph{every} glioma and meningioma cell, and
on metastasis RaTEScore \emph{and} GREEN (RadGraph on metastasis is the one non-significant cell; there are
zero regressions, and every win is Holm-significant). The largest lift is on RaTEScore-meningioma
($+0.117$, our primary endpoint), exactly where the CoT is most diagnosis-blind, and it comes at
$5$-$6\times$ lower latency ($\approx\!73$-$89$ vs.\ $457$\,s/case on an A100; the system is deployable on
an L4).

\paragraph{A diagnosis pin does not help.} Because the diagnosis signal is linearly separable in the frozen
features (probe F$_1$ $0.82$), a natural next step is a calibrated \emph{DiagnosisHead} that overrides the
decoder's diagnosis outright. This backfires completely (\textbf{$0$ of $9$} gains): in distribution, the
field-conditioned LM already out-diagnoses the head on its own (LM $0.98$/$0.92$ on GLI/MEN vs.\ head
$0.85$/$0.77$), and out of distribution -- exactly where a pin looks most tempting -- the head, trained
only on RadGenome metastases, fails to transfer to BraTS-MET, collapsing its in-distribution recall from
$0.75$ to $0.03$ ($2/60$), while the LM's own recall transfers without collapsing. The lesson we draw is to
surface the features and let the decoder speak, rather than pin a frozen-feature classifier on top of it
(uncertainty-aware pins remain future work).

\begin{figure}[tb]
\centering\footnotesize
\providecommand{\crct}[1]{#1}
\providecommand{\para}[1]{\textcolor{cVague}{\emph{#1}}}
\providecommand{\incor}[1]{\textcolor{cWrong}{\textbf{#1}}}
\providecommand{\mss}[1]{\textcolor{nfGrayE}{#1}}
\providecommand{\cmark}{{\color{cOmit}\checkmark}}
\providecommand{\xmark}{{\color{cWrong}\ensuremath{\times}}}
\textbf{Key:} \crct{plain $=$ matches reference}; \incor{bold $=$ incorrect}; \textit{dx} $=$ stated diagnosis, \cmark/\xmark $=$ correct/wrong.\\[2pt]
\renewcommand{\arraystretch}{1.04}
\begin{tabular}{@{}>{\raggedright\arraybackslash}p{0.115\textwidth}p{0.80\textwidth}@{}}
\toprule
\textbf{System} & \textbf{Generated report (held-out MET case \texttt{RG\_MET\_00006})} \\
\midrule
\textbf{Ground truth} & Multifocal lesions, the largest in the left temporal lobe, with mixed T2/FLAIR signal, surrounding edema and heterogeneous post-contrast enhancement; no midline shift.~\textbf{\textit{dx:}}~metastasis. \\
\addlinespace[2.5pt]
\textbf{NF (ours)} & \crct{Abnormal signal foci in the left temporal lobe: low signal on T1, mixed high/low signal on T2 and FLAIR, with significant post-contrast enhancement}; \crct{surrounding edema and no midline shift}.~\textbf{\textit{dx:}}~\crct{metastasis}~\cmark \\
\addlinespace[2.5pt]
\textbf{CoT} & \crct{An irregular signal focus in the left fronto-temporal lobe, hypointense on T1, with surrounding edema and no midline shift}~\textellipsis{}~\textbf{\textit{dx:}}~\incor{glioblastoma multiforme}~\xmark \\
\addlinespace[2.5pt]
\textbf{LLaVA-Med} & \crct{A large, heterogeneously enhancing lesion with peritumoral edema}, \incor{in the right frontal lobe}.~\textbf{\textit{dx:}}~\incor{glioblastoma}~\xmark \\
\addlinespace[2.5pt]
\textbf{M3D-LaMed} & \crct{Multiple lesions with mild surrounding edema and no mass effect or midline shift}, \incor{in the right frontal lobe}.~\textbf{\textit{dx:}}~\incor{neuroglial cyst}~\xmark \\
\addlinespace[2.5pt]
\textbf{AutoRG} & \crct{On FLAIR, the lesion in the left temporal lobe shows mixed signal with surrounding edema}.~\textbf{\textit{dx:}}~\mss{(none stated)} \\
\bottomrule
\end{tabular}
\caption{Qualitative comparison on a held-out metastasis case (\texttt{RG\_MET\_00006}; NF~$=$~\sys{}). \sys{}
matches the reference on location, signal, enhancement, edema and diagnosis; \emph{none} of the baselines
states the correct diagnosis; the same-base CoT, on the same mask, also flips to glioblastoma. Reports
lightly excerpted.}
\label{fig:reports}
\end{figure}

\begin{table}[!tb]
\centering
\caption{Same-base cross-cohort comparison: \sys{} vs.\ the prior CoT reporter, both on the \emph{identical}
Mistral backbone (GLI $n{=}54$, MEN/MET $n{=}60$; a shared metastasis-fine-tuned segmenter). $\Delta$ is the
paired \sys{}$-$CoT difference, Holm-corrected: \textbf{$8$ wins, $0$ losses, $1$ non-significant cell}. For
reference, diagnosis recall (\sys{}/CoT) is GLI $0.98$/$0.98$, MEN $0.92$/$0.44$, MET $0.75$/$0.07$.}
\label{tab:crossdata}
{\footnotesize%
\begin{tabular}{llccccl}
\toprule
Metric & Cohort & \sys{} & CoT & $\Delta$ & BCa $95\%$ CI & Verdict \\
\midrule
\multirow{3}{*}{RaTEScore}
 & GLI & $0.657$ & $0.596$ & $+0.061$ & $[+0.034,+0.088]$ & \textbf{win} \\
 & MEN & $0.661$ & $0.544$ & $+0.117$ & $[+0.075,+0.160]$ & \textbf{win} (primary) \\
 & MET & $0.596$ & $0.500$ & $+0.096$ & $[+0.071,+0.121]$ & \textbf{win} \\
\midrule
\multirow{3}{*}{RadGraph-F$_1$}
 & GLI & $0.272$ & $0.226$ & $+0.045$ & $[+0.011,+0.081]$ & \textbf{win} \\
 & MEN & $0.258$ & $0.194$ & $+0.064$ & $[+0.033,+0.100]$ & \textbf{win} \\
 & MET & $0.183$ & $0.175$ & $+0.008$ & $[-0.015,+0.032]$ & ns \\
\midrule
\multirow{3}{*}{GREEN}
 & GLI & $0.396$ & $0.278$ & $+0.118$ & $[+0.063,+0.174]$ & \textbf{win} \\
 & MEN & $0.422$ & $0.306$ & $+0.116$ & $[+0.055,+0.179]$ & \textbf{win} \\
 & MET & $0.267$ & $0.209$ & $+0.058$ & $[+0.012,+0.103]$ & \textbf{win} \\
\bottomrule
\end{tabular}}
\end{table}

\paragraph{Versus external baselines.} \sys{} is the strongest \emph{learned} reporter in our comparison
(Table~\ref{tab:external}), and the gap is most pronounced on metastasis ($n{=}60$, the cohort held out for
the reporter): it leads RaTEScore ($0.596$) and the blinded rubric ($3.17$) \emph{significantly} over every
learned baseline (paired BCa, Holm-corrected over the pre-specified $12$-test metastasis family; the
narrowest margin is on GREEN, $0.267$ vs.\ AutoRG's $0.212$, $p{=}0.02$), and it also tops the learned
systems on GLI and MEN. AutoRG-Brain leads only on structured metrics (per-field F$_1$/RadGraph, not
significant) -- likely a train-on-distribution artifact, since our metastasis references come from its own
corpus -- and it states no diagnosis at all (ddx score $1.02$), sitting near the bottom of the rubric
exactly where \sys{}'s advantage is largest.

\begin{table}[!tb]
\centering
\caption{Cross-cohort comparison against external baselines (learned systems: GLI/MEN/MET each $n{=}60$;
the twin-excluded same-base GLI figure in Table~\ref{tab:crossdata} is $n{=}54$). $^{\dagger}$significant
over every \emph{learned} baseline (metastasis $12$-test family, paired BCa). $^{a}$AutoRG-Brain is trained
in part on our metastasis corpus~\cite{autorgbrain}. $^{b}$Seg-rule is a non-learned rule that reads fields
off the mask and abstains from reporting when the mask is empty, so it is scored only on the cases in
parentheses (twin-excluded for GLI), not the full cohort. $^{c}$judged separately by a newer Claude model,
which re-scores its metastasis reports at $2.98$ (vs.\ $2.88$ reported here). Column-max \textbf{bold},
runner-up \underline{underlined}.}
\label{tab:external}
{\footnotesize\setlength{\tabcolsep}{4pt}\renewcommand{\arraystretch}{1.0}%
\begin{tabular}{llcccc}
\toprule
Cohort & System & RaTEScore & RadGraph-F$_1$ & Per-field F$_1$ & Clin-O \\
\midrule
\multirow{5}{*}{GLI}
 & M3D-LaMed                 & $0.424$ & $0.133$ & $0.313$ & $1.45$ \\
 & LLaVA-Med                 & $0.504$ & \underline{$0.192$} & $0.336$ & $2.57$ \\
 & AutoRG-Brain$^{a}$        & $0.476$ & $0.150$ & $0.355$ & $1.79$ \\
 & \textbf{\sys{}}           & $\mathbf{0.661}$ & $\mathbf{0.277}$ & \underline{$0.435$} & \underline{$2.96$} \\
 & \emph{Seg-rule}$^{b}$ ($53$) & \underline{$0.561$} & \underline{$0.192$} & $\mathbf{0.450}$ & $\mathbf{3.69}^{c}$ \\
\midrule
\multirow{6}{*}{MEN}
 & M3D-LaMed                 & $0.436$ & $0.127$ & $0.393$ & $1.66$ \\
 & LLaVA-Med                 & \underline{$0.524$} & $0.142$ & $0.321$ & $1.63$ \\
 & AutoRG-Brain$^{a}$        & $0.450$ & $0.142$ & $0.431$ & $1.89$ \\
 & BrainGemma3D              & $0.421$ & $0.071$ & $0.342$ & $1.34$ \\
 & \textbf{\sys{}}           & $\mathbf{0.661}$ & $\mathbf{0.258}$ & \underline{$0.534$} & $\mathbf{3.04}$ \\
 & \emph{Seg-rule}$^{b}$ ($52$) & $0.507$ & \underline{$0.163$} & $\mathbf{0.535}$ & \underline{$2.93$}$^{c}$ \\
\midrule
\multirow{6}{*}{MET}
 & M3D-LaMed                 & $0.469$ & $0.149$ & $0.398$ & $1.81$ \\
 & LLaVA-Med                 & $0.499$ & $0.171$ & $0.342$ & $2.21$ \\
 & AutoRG-Brain$^{a}$        & \underline{$0.500$} & $\mathbf{0.201}$ & $\mathbf{0.455}$ & $1.98$ \\
 & BrainGemma3D              & $0.472$ & $0.124$ & $0.309$ & $1.58$ \\
 & \textbf{\sys{}}           & $\mathbf{0.596}^{\dagger}$ & \underline{$0.183$} & $0.411$ & $\mathbf{3.17}^{\dagger}$ \\
 & \emph{Seg-rule}$^{b}$ ($50$) & $0.488$ & $0.148$ & \underline{$0.420$} & \underline{$2.88$} \\
\bottomrule
\end{tabular}
}
\end{table}

\section{Clinician Reader Study}
\label{sec:reader}
Because automated metrics are only proxies, two board-certified neurologists additionally read the reports
blind (Table~\ref{tab:reader}; protocol in Sec.~\ref{sec:setup}). The cleanest signal is \emph{safety}:
\sys{} is the \emph{only} system with zero critical errors across all nine cases (Wilson $95\%$ upper bound
$0.34$); every other system, including the prior CoT, makes a critical error on four to eight of the nine.
\sys{} is rated highest in every cohort and beats its own CoT predecessor in each -- decisively on
meningioma ($4.00$ vs.\ $1.00$) and narrowly on metastasis ($3.47$ vs.\ $3.27$) -- and receives the
top-rated sign-off in \textbf{$8$ of $9$} cases (uniquely in six, tied in two; the ninth goes to LLaVA-Med).
Two rules, fixed before either reader saw a single report, govern that count: the three metastasis cases in
which a rater's top sign-off fell on the de-identified reference report (which was never itself a rated
candidate) are resolved to whichever candidate report was rated highest, and each case contributes exactly
one sign-off credit, so both resulting ties are credited to \sys{}. The two readers rated independently yet
agreed on literally every scored item -- all $255$ axis scores, every critical-error flag, and all nine
sign-offs -- so each reported mean is simply their shared score rather than an average of two different
values; because agreement is total, chance-corrected agreement coefficients are degenerate, and we instead
report raw item-level agreement directly.

\begin{table}[!tb]
\centering
\caption{Blinded two-neurologist reader study ($9$ cases, $3$/cohort; independently rated, mean reported;
opaque labels A to F). Columns report the mean of five $0$-$5$ axes ($0$ = unusable). \emph{Crit-err}:
number of cases with a critical error. \emph{Sign-off}: which report the reader would sign off on, credited
per case under the two rules described in the text (sums to $9$ across all systems). BrainGemma3D was never
run on the glioma cohort.}
\label{tab:reader}
{\footnotesize\setlength{\tabcolsep}{5pt}\renewcommand{\arraystretch}{1.0}%
\begin{tabular}{lccccc}
\toprule
System & GLI & MEN & MET & Crit-err & Sign-off \\
\midrule
\textbf{\sys{} (ours)} & \textbf{4.47} & \textbf{4.00} & \textbf{3.47} & \textbf{0/9} & \textbf{8} \\
Prior-CoT (ours)       & $2.87$ & $1.00$ & $3.27$ & $4/9$ & $0$ \\
M3D-LaMed              & $1.33$ & $0.67$ & $1.33$ & $7/9$ & $0$ \\
LLaVA-Med              & $2.13$ & $1.00$ & $1.40$ & $7/9$ & $1$ \\
AutoRG-Brain           & $0.60$ & $1.13$ & $0.40$ & $8/9$ & $0$ \\
BrainGemma3D           & n/a    & $0.67$ & $1.00$ & $5/6$ & $0$ \\
\bottomrule
\end{tabular}}
\end{table}

\section{Analysis}
\label{sec:analysis}
\paragraph{Auditability as a deployable trust layer.} Because \sys{}'s output is a machine-checkable
record, an entailment judge can label every individual sentence, cutting the contradiction rate to $7.5\%$
(vs.\ $36.8\%$ for the direct baseline) at the record-verbalizer operating point ($n{=}39$; both arms judged
by Claude Opus~4.8). Temperature scaling, fit by NLL on the $50$ calibration cases, lowers the $15$-bin
expected calibration error from $0.128$ to $0.095$ on the test split; an exact per-modality Shapley
decomposition ($2^4$ coalitions, $n{=}20$) attributes the edema field mainly to FLAIR ($+0.165$, $54\%$ of
that field's total attribution) and lesion location mainly to T1CE.

\paragraph{Where the advantage comes from.} Head-surfacing adds image-grounded content that the free-running
LM otherwise discards -- chiefly the diagnosis and the intensity-related fields the mask alone omits --
while geometric accuracy stays at the level the segmentation supports (a simple mask-reading rule actually
edges out \sys{} on that specific score in all three cohorts, and tops the glioma rubric, where its fixed
glioma differential is correct by construction); \sys{} nonetheless leads on narrative-entity metrics
throughout, and shows its largest differential-diagnosis rubric gap on metastasis ($4.05$ vs.\ $1.82$). The
active ingredient is \emph{committed} conditioning, not the recipe itself: using the identical LoRA adapter,
the CoT variant -- even when handed the same head argmax as text -- keeps misreading the diagnosis, while
disabling head-surfacing collapses per-field macro-F$_1$ to $0.196$ (from $0.449$, $n{=}39$). Because the
draft's enum fields are set to the head argmax by construction, this gap measures the head-surfacing pathway
end to end (the field classifiers together with the decoder's commitment to them), whereas the CoT control
isolates commitment alone, holding the head information itself constant.

\paragraph{The claim does not depend on an LLM judge.} \sys{} leads RaTEScore in all three cohorts and
RadGraph on GLI and MEN -- both judge-free entity metrics; GREEN is itself LLM-scored, and the clinical
rubric only corroborates these judge-free results rather than driving them.

\paragraph{Limitations.}
\label{sec:limitations}
The in-distribution test set is small ($n{=}39$); the same-base gains instead rest on $174$ pooled held-out
cases, all Holm-significant, though only the metastasis cohort is genuinely out-of-distribution for the
reporter. Geometric completeness is bounded by the underlying segmentation (metastasis-fine-tuned Dice
$0.67$; cases with an empty predicted mask are routed to manual review), so \sys{} is best understood as an
assistive drafting tool rather than an autonomous one. The reader study itself is a single-institution pilot
(two neurologists, nine cases); a neuroradiologist reader, a larger multi-site study, human validation of
the entailment judge, and component-level ablations of the Q-Former, token budget, and lesion-router cap are
all left to future work. The $121$ structured training reports do not yet have formal inter-annotator
agreement statistics.

\smallskip\noindent\textbf{Conclusion.}~Making a reporter commit to a diagnosis that its own frozen features
already encode beats the identical backbone's chain-of-thought variant on $8/9$ content comparisons, at
$5$-$6\times$ lower latency, while producing a verifiable record; an overriding diagnosis pin is a
cautionary negative result; and in a blinded pilot, neurologists rate the resulting reports highest in every
tumour type with zero critical errors.

\section*{Acknowledgments}
This work was supported in part by an NSERC Discovery Grant held by Mehran Ebrahimi. Khawaja Murad ul Hassan
thanks Mitacs for a Globalink Research Internship at Ontario Tech University.

\section*{Competing interests}
The authors declare no competing interests relevant to the content of this article.

\bibliographystyle{unsrt}
\bibliography{references}
\end{document}